\documentclass[letterpaper]{article} 
\usepackage[preprint]{aaai2027}  
\usepackage[hyphens]{url}  
\usepackage{graphicx} 
\usepackage{natbib}  
\usepackage{caption} 
\usepackage{algorithm}
\usepackage{algorithmic}

\usepackage{newfloat}
\usepackage{listings}
\DeclareCaptionStyle{ruled}{labelfont=normalfont,labelsep=colon,strut=off} 
\floatstyle{ruled}
\newfloat{listing}{tb}{lst}{}
\floatname{listing}{Listing}

\usepackage{booktabs}
\usepackage{amsmath}
\usepackage{amssymb}
\usepackage{siunitx}
\usepackage{colortbl}
\definecolor{tableOurs}{RGB}{236,243,250}
\usepackage{multirow} 
\title{Pixel-Space Diffusion via Observation Operators}
\author{
    Shaojie Guo\textsuperscript{\rm 1},
    Lichen Ma\textsuperscript{\rm 2,\rm 3},
    Haoyang Tong\textsuperscript{\rm 2},
    Yu He\textsuperscript{\rm 2},
    Zipeng Guo\textsuperscript{\rm 2},
    Xiaoan Liu\textsuperscript{\rm 2},
    Feng Yan\textsuperscript{\rm 2},
    Yu Guo\textsuperscript{\rm 2},
    Fei Wang\textsuperscript{\rm 2},
    Junshi Huang\textsuperscript{\rm 2},
    Yan Wang\textsuperscript{\rm 1}
}
\affiliations{
    \textsuperscript{\rm 1}East China Normal University\\
    \textsuperscript{\rm 2}JD.com\\
    \textsuperscript{\rm 3}State Key Laboratory of Human-Machine Hybrid Augmented Intelligence, Institute of Artificial Intelligence and Robotics, Xi'an Jiaotong University\\
}

\begin{document}

\maketitle

\begin{abstract}
Pixel-space diffusion models directly model image distributions but remain difficult to optimize.
Recent methods alleviate this challenge through target reparameterization, while still relying on a fixed clean-image target throughout denoising.
Through empirical analysis, we identify a scale--time mismatch: image structures become predictable from coarse to fine as noise decreases, whereas existing models are forced to predict the full image even under high noise, resulting in low-SNR gradients that hinder optimization.
To resolve this mismatch, we propose Observation Operator Diffusion, a unified framework that aligns both the supervision trajectory and feature refinement with the intrinsic recovery order of image structures.
Specifically, we replace fixed full-image supervision along the standard flow path with a time-indexed observation trajectory that evolves from coarse structures to the full image during denoising.
This trajectory is instantiated with a family of Gaussian--Lanczos operators at varying observation scales, yielding a path-consistent training objective.
We further introduce GL-CoDA, a decoder that injects scale-specific Gaussian--Lanczos observations across decoding stages for coarse-to-fine feature refinement. 
Extensive experiments show that the proposed approach converges substantially faster while consistently improving generation quality, achieving an FID of 1.52 on ImageNet-256.
\end{abstract}

\section{Introduction}

Pixel-space diffusion has recently regained attention~\citep{li2025back}. Unlike latent diffusion~\citep{rombach2022high}, which generates within a compressed representation, it models images directly in pixel space. This avoids information loss from autoencoder compression and reconstruction errors introduced by a fixed decoder, providing a direct route to high-fidelity generation~\citep{chen2025pixelflow,yu2026pixeldit}. However, this advantage comes with substantially greater optimization difficulty~\citep{chen2026asymflow}.

\begin{figure}[t]
    \centering
    \includegraphics[width=\columnwidth]{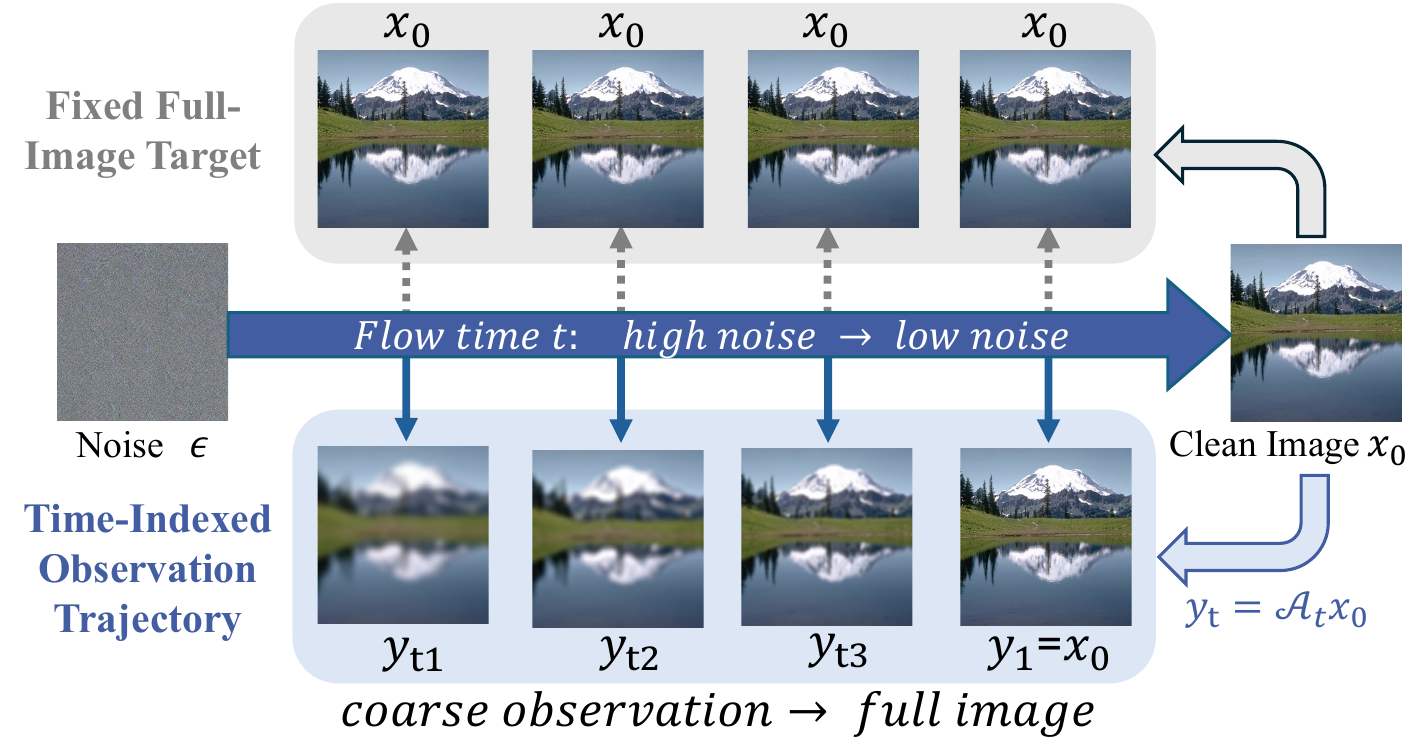}
    \caption{\textbf{Fixed full-image supervision vs.\ time-indexed observation trajectory.}
Existing pixel-space diffusion models use the full clean image $x_0$ as the target throughout the denoising path. We instead introduce a time-indexed observation trajectory $y_t=\mathcal{A}_t(x_0)$, whose image content evolves from coarse structure to the full image as denoising proceeds.}
    \label{fig:neural}
    \vspace{-5mm}
\end{figure}

Recent studies show that optimization depends critically on the prediction target: even at a fixed pixel dimensionality, target reparameterization can markedly accelerate convergence~\citep{li2025back}, motivating increasing attention to target design~\citep{chen2026asymflow,jin2026revisiting}. 
However, these methods change only the target parameterization while retaining the full image supervision throughout denoising.
This distinction is particularly important in pixel space. 
Unlike latent-space models, where a compressed target delegates image reconstruction to a decoder, pixel-space models must directly predict image structures spanning multiple scales, from global layout to local textures. 
Yet the recoverability of these structures changes continuously with noise level: coarse structures emerge earlier, while fine details become predictable only as noise decreases (Fig.~\ref{fig:neural}). 
Full-image supervision therefore forces the model to fit unrecoverable fine details at early denoising stages. 
We term this discrepancy between the fixed supervised scale and the time-dependent recoverable scale a scale--time mismatch, and identify it as a fundamental yet under-recognized source of optimization difficulty in pixel-space diffusion.

\begin{figure}[t]
    \centering
    \includegraphics[width=\columnwidth]{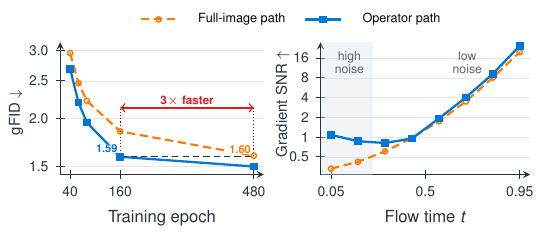}
            \vspace{-5mm}
\caption{\textbf{Optimization benefits of the observation trajectory.}
\textbf{Left:} Observation Operator Diffusion converges substantially faster than conventional full-image supervision while achieving a lower final FID.
\textbf{Right:} Gradient signal-to-noise ratio (SNR) over diffusion time. By avoiding supervision of unresolved fine details, the observation trajectory maintains higher-SNR gradients in the high-noise regime, leading to more efficient optimization.}
        \vspace{-3mm}
    \label{fig:fid}
\end{figure}

To substantiate this claim, we measure prediction errors across spatial scales throughout denoising (Fig.~\ref{fig:scale-predictability}; measurement protocol in the Appendix). 
The results reveal a clear coarse-to-fine recovery order: under high noise, coarse structures can already be reliably inferred, while fine details remain ambiguous and become recoverable only as noise decreases. 
This behavior is misaligned with the fixed full-image supervision used by existing pixel-space models, which requires the network to predict uncertain fine details even at high noise levels. Such a scale--time mismatch introduces low-SNR supervision signals, reducing the efficiency of optimization and ultimately slowing convergence (Fig.~\ref{fig:fid}).

\begin{figure}[t]
    \centering
    \includegraphics[width=0.9\columnwidth]{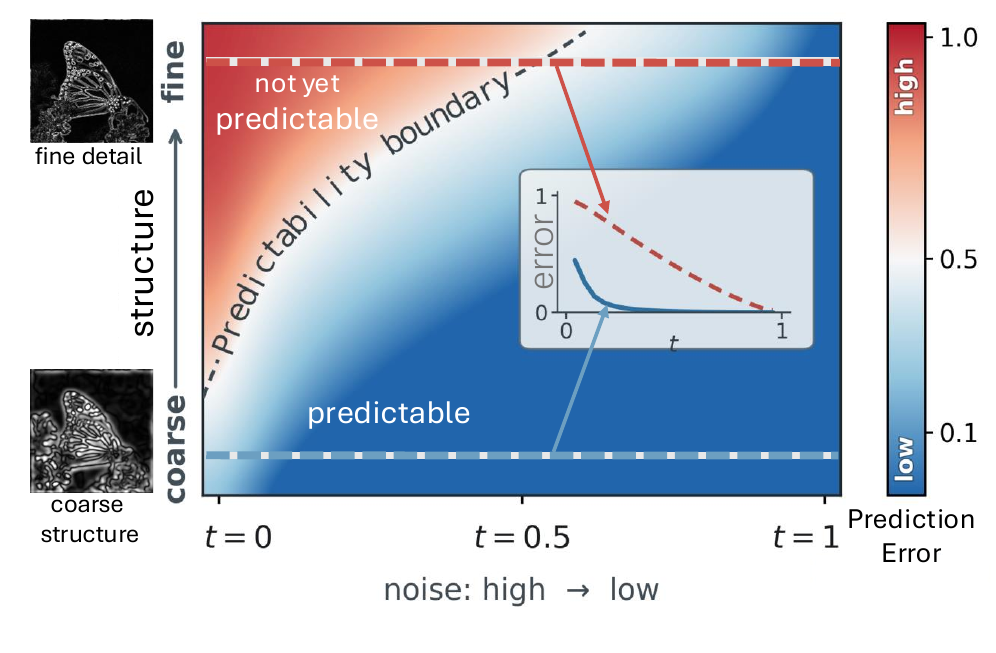}
        \vspace{-5mm}
    \caption{\textbf{Scale-dependent predictability during denoising.}
    We decompose the single-step prediction residual into multiscale structural bands and measure the prediction error at each scale. The heatmap (denoising time $\times$ spatial scale) and representative trajectories (inset) show that coarse structures become recoverable earlier than fine details, exposing the mismatch with fixed full-image supervision.}
    \label{fig:scale-predictability}
        \vspace{-5mm}
\end{figure}

A natural solution is to supervise pixel-space models only on the image structures that are predictable at the current noise level, rather than on the full image throughout denoising. 
The prediction target should therefore evolve with flow time, following the coarse-to-fine order in which image structures become recoverable. 
Building on this idea, we propose Observation Operator Diffusion, which replaces the fixed full-image target in the standard flow path with a time-indexed observation trajectory. 
We instantiate this trajectory with separable Gaussian--Lanczos operators, whose controllable scale selectivity enables progressive coarse-to-fine supervision while suppressing ringing artifacts. 
By annealing their scales over time, the observations evolve smoothly from a coarse low-pass image to the full image. 
The existing full image path can be considered a special case when the observation operator is an identity operator.

We further derive a path-consistent objective to explicitly couple the supervision scale with flow time, ensuring that each noisy input is trained against only the image structures recoverable at its current noise level. 
At high noise levels, this prevents unresolved fine details from interfering with the learning target, producing higher-SNR gradients and accelerating convergence (Fig.~\ref{fig:fid}). 
This coarse-to-fine principle is also extended from the evolution of supervision over time to the refinement of features across network depth. 
To this end, we introduce GL-CoDA, a Gaussian--Lanczos Coarse-to-Detail Aggregation decoder that injects scale-specific structural responses into decoding stages and progressively aggregates them from coarse to fine. 
In this way, the decoder follows a progressive refinement of the representation from the global structure to fine-grained details.
Our contributions are summarized as follows:
\begin{itemize}
\item We reveal an overlooked source of optimization difficulty in pixel-space diffusion, termed \emph{scale--time mismatch}, arising from the discrepancy between the coarse-to-fine structure recovery and fixed full-image supervision.

\item We propose Observation Operator Diffusion, which replaces fixed full-image supervision with a time-indexed trajectory of Gaussian--Lanczos observations, and derive a path-consistent objective that aligns the supervised scale with the structure recoverable at each denoising step.

\item We introduce GL-CoDA, a scale-ordered decoder that injects Gaussian--Lanczos  responses across decoding stages, extending the coarse-to-fine principle to feature refinement over network depth.

\item Extensive experiments demonstrate substantially faster convergence and consistently improved generation quality, achieving an FID of \textbf{1.52} on ImageNet-256.
\end{itemize}

\section{Related Work}

\paragraph{Pixel-Space Diffusion and Prediction Objectives.}
Pixel-space diffusion avoids the reconstruction bottleneck of latent autoencoders, but remains challenging to optimize~\citep{ho2020denoising,dhariwal2021diffusion,peebles2023scalable}.
Recent pixel-space architectures improve this tradeoff through patch--pixel modeling, hyper-connected feature reuse, or U-shaped transformer designs~\citep{yu2026pixeldit,he2026hyperdit,guo2026pixelu}.
Recent studies have explored alternative prediction objectives to improve optimization. 
JiT adopts clean-image prediction to exploit the low-dimensional structure of natural images~\citep{li2025back}; 
Asymmetric Flow Models restrict noise prediction to a low-rank subspace while retaining full-dimensional data prediction~\citep{chen2026asymflow}; 
and k-Diff adapts the prediction target according to data dimensionality~\citep{jin2026revisiting}. 
Despite different parameterizations, they all retain full-image supervision throughout denoising. 
Our work instead considers the temporal evolution of structure recoverability and adapts the supervision scale accordingly, following the intrinsic coarse-to-fine recovery order of image structures.

\paragraph{Scale-Aware Pixel Diffusion.}
Several recent methods incorporate the multiscale nature of image generation into pixel-space diffusion.
DeCo~\citep{ma2026deco} assigns low-frequency semantics and high-frequency details to different network components and introduces a frequency-aware loss.
FREPix~\citep{frepix} decomposes generation into low- and high-frequency components with separate transport paths.
Spectral Forcing~\citep{spectralforcing} filters noisy inputs in the DCT domain to suppress high-frequency components dominated by noise.
FrequencyBooster~\citep{ma2026frequencybooster} further strengthens pixel-space modeling with a high-capacity decoder for full-frequency detail recovery.
Energy-Guided Flow Matching~\citep{tong2026egfm} replaces the fixed clean endpoint with an energy-scheduled moving endpoint to explicitly encode coarse-to-fine generation.
These works highlight the importance of frequency and scale structure, while our method reformulates the supervision target itself as a continuous observation trajectory whose scale is explicitly coupled to diffusion time.

\paragraph{Scale-Space Representations and Image Operators.}
Scale-space theory models an image as a family of continuously indexed observations at different spatial scales~\citep{witkin1983scale,lindeberg1994scale}. 
Related image operators, including Gaussian filters, Laplacian pyramids, and Lanczos kernels, have long been used to represent, analyze, and resample image content across spatial scales~\citep{burt1983laplacian,turkowski1990filters}.
The concept of using scale as an image representation attribute has been initially explored in diffusion models~\citep{hoogeboom2022blurring}.
Our method establishes this connection by constructing a diffusion-time-indexed family of Gaussian--Lanczos observations, turning spatial scale into a dynamic prediction target that evolves continuously throughout denoising.

\begin{figure*}[t]
    \centering
    \includegraphics[width=\textwidth]{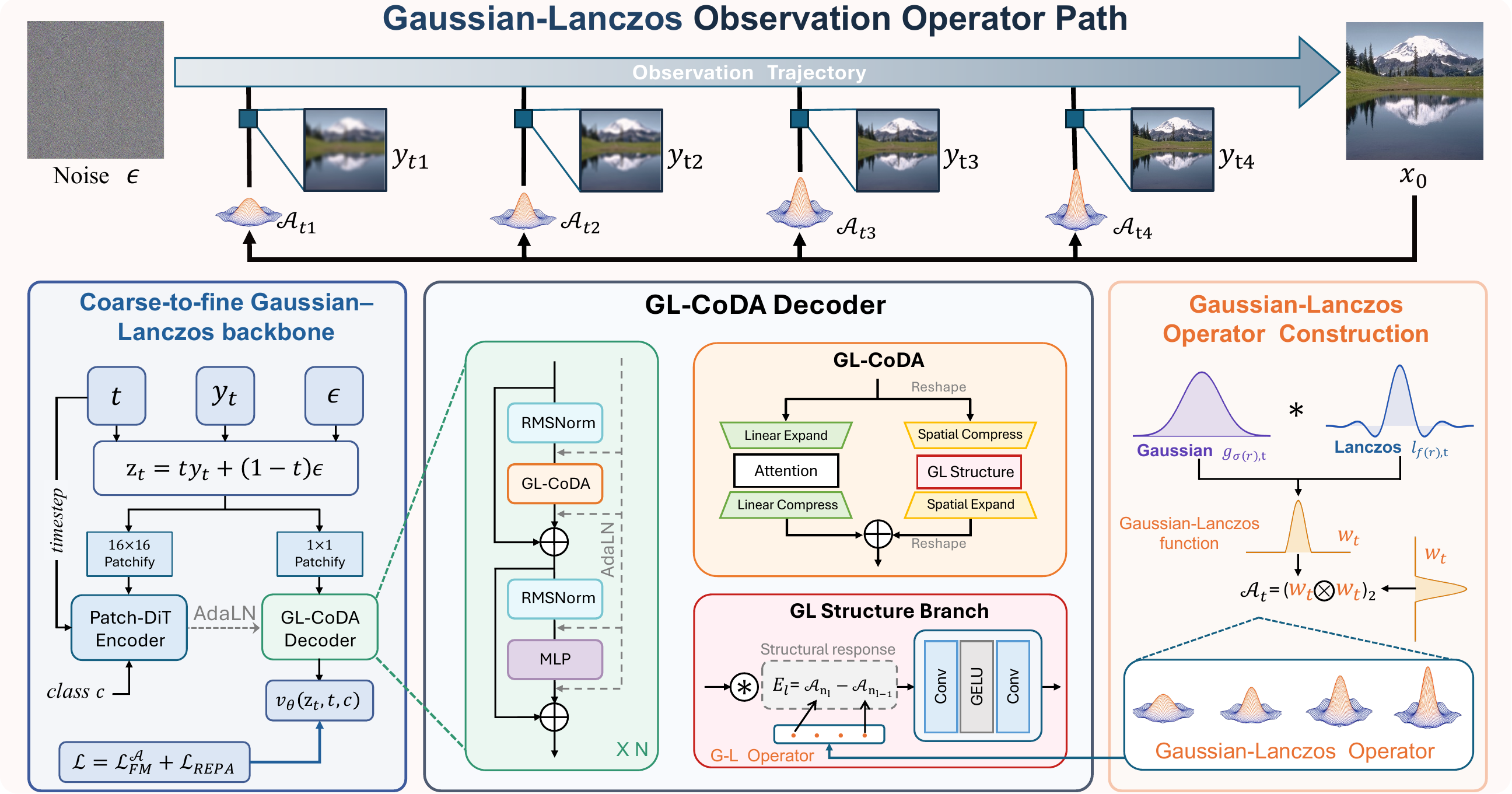}
\caption{\textbf{Overview of the proposed Observation Operator Diffusion framework.}
\textbf{Top:} Gaussian--Lanczos observation trajectory, where a family of operators $\mathcal{A}_t$ maps the clean image $x_0$ to time-indexed observations $y_t=\mathcal{A}_t(x_0)$, progressively transitioning from coarse structures to the full image along denoising.
\textbf{Bottom:} two-stage Patch--Pixel backbone, comprising a Patch-DiT encoder and a GL-CoDA decoder. The GL structure branch injects scale-specific structural responses across decoding stages. The right panel details the construction of each Gaussian--Lanczos operator $\mathcal{A}_t$.}
    \label{fig:method-framework}
        \vspace{-3mm}
\end{figure*}

\section{Method}
\label{sec:method}
Figure~\ref{fig:method-framework} presents the proposed observation operator diffusion framework.
Motivated by the empirically observed coarse-to-fine predictability during denoising, we replace the fixed full-image target with a time-indexed Gaussian--Lanczos observation trajectory and derive a path-consistent training objective.
To extend the same coarse-to-fine principle to feature refinement, we further introduce a two-stage Patch--Pixel backbone with a GL-CoDA decoder, whose GL structure branch injects scale-specific structural responses across decoding stages.
By jointly aligning supervision over denoising time and feature refinement over network depth, the proposed framework improves both optimization efficiency and generation quality.

\subsection{Scale--Time Mismatch in Pixel-Space Diffusion}
\label{sec:prelim}

Recent pixel-space diffusion models are commonly trained using flow
matching~\citep{lipman2023flow}.
Given a clean image $x_0\sim p_{\mathrm{data}}$ and Gaussian noise
$\epsilon\sim\mathcal N(0,I)$, flow matching defines the linear path
\begin{equation}
z_t = t x_0 + (1-t)\epsilon,
\qquad t\in(0,1),
\label{eq:linear-fm}
\end{equation}
with target velocity
$v_t^\star=\dot z_t=x_0-\epsilon$.
At each training iteration, the model takes the noisy state $z_t$,
flow time $t$, and class condition $c$ as input.
For a prediction parameterization
$\pi\in\{x,\epsilon,v\}$, the model produces
\begin{equation}
r_\theta^\pi
=
F_\theta^\pi(z_t,t,c).
\label{eq:direct-prediction}
\end{equation}

Although these parameterizations predict different quantities, their
prediction errors can be mapped to a common image space.
Following the unified formulation of JiT~\citep{li2025back}, each direct
prediction induces an estimate of the clean image:
\begin{equation}
\hat{x}_\theta^\pi =
\begin{cases}
r_\theta^x,
& \pi=x, \\[2pt]
\dfrac{z_t-(1-t)r_\theta^\epsilon}{t},
& \pi=\epsilon, \\[7pt]
z_t+(1-t)r_\theta^v,
& \pi=v,
\end{cases}
\qquad
\pi\in\{x,\epsilon,v\}.
\label{eq:implied-clean-prediction}
\end{equation}
Let
$\delta x_\theta^\pi=\hat{x}_\theta^\pi-x_0$
denote the corresponding image-space error.
The standard mean-squared-error objective can be expressed
in terms of this common image-space error as
\begin{equation}
\mathcal L_\pi
=
\mathbb E\!\left[
w_\pi(t)
\left\|\delta x_\theta^\pi\right\|_2^2
\right],
\qquad
\pi\in\{x,\epsilon,v\},
\label{eq:prediction-loss-parameterization}
\end{equation}
where $w_x(t)=1$, $w_\epsilon(t)=t^2/(1-t)^2$, and
$w_v(t)=1/(1-t)^2$.
Equations~\eqref{eq:implied-clean-prediction}--\eqref{eq:prediction-loss-parameterization}
show that, in the common image space, target reparameterization only
introduces a timestep-dependent scalar weighting on the prediction error.
Although different
parameterizations emphasize different timesteps, they all evaluate the induced clean-image
estimate against the same full-resolution target $x_0$.
Since a scalar rescales the entire error uniformly, coarse structure and fine detail remain
equally present in the training signal at every timestep.
 
This fixed supervision is misaligned with the coarse-to-fine order in
which image structures become recoverable during denoising.
As shown in Fig.~\ref{fig:scale-predictability}, coarse structures can be
reliably recovered well before fine details.
Consequently, full-image supervision requires the model to fit fine
details before they are reliably inferable from the noisy input, introducing low-SNR gradients at high noise levels.
This scale--time mismatch reduces optimization efficiency, slowing convergence and ultimately degrading generation quality
(Fig.~\ref{fig:fid}).

\subsection{Gaussian-Lanczos Observation Operator Path}
\label{sec:scale-prediction}

To align the spatial content of supervision with the image structures
that are reliably predictable at each flow time, we replace the fixed
full-image target with a time-indexed observation trajectory.
Specifically, we introduce a family of operators
$\{
\mathcal A_t:
\mathbb R^{C\times H\times W}
\rightarrow
\mathbb R^{C\times H\times W}
\}_{t\in[0,1]}$
and define
\begin{equation}
y_t=\mathcal A_t(x_0).
\label{eq:scale-observation}
\end{equation}

Each operator preserves the spatial resolution of the image while
controlling the spatial scale of the observed content.
We require the operator family to satisfy
\begin{equation}
\mathcal A_0=\mathcal A_{\mathrm{coarse}},
\qquad
\mathcal A_1=\mathcal I,
\label{eq:observation-endpoints}
\end{equation}
where $\mathcal A_{\mathrm{coarse}}$ produces a coarse low-pass
observation and $\mathcal I$ denotes the identity operator.
Thus, the observation trajectory starts from coarse image structure and
gradually recovers finer details, reaching the full clean image at
$t=1$:
$y_0=\mathcal A_{\mathrm{coarse}}(x_0)$ and $y_1=x_0$.
To ensure that this trajectory is compatible with flow matching, we
further require $t\mapsto\mathcal A_t(x)$ to be absolutely continuous
for every image $x$.
The resulting observation trajectory is therefore continuous and
differentiable almost everywhere, so that its flow-matching velocity is
well defined.
Using $y_t$ as a time-dependent prediction target we construct the
observation path
\begin{equation}
z_t
=
t\,y_t+(1-t)\epsilon
=
t\,\mathcal A_t(x_0)+(1-t)\epsilon.
\label{eq:scale-aware-path}
\end{equation}
The path starts from pure noise, $z_0=\epsilon$, and terminates at the
clean image, $z_1=y_1=x_0$.
Differentiating Eq.~\eqref{eq:scale-aware-path} with respect to flow
time gives the path-consistent velocity target
\begin{equation}
v_t^\star
=
\dot z_t
=
y_t-\epsilon+t\,\dot y_t,
\qquad
\dot y_t
=
\partial_t\mathcal A_t(x_0).
\label{eq:scale-aware-velocity}
\end{equation}
Compared with the standard flow-matching target, the additional term
$t\dot y_t$ captures the temporal evolution of the observation anchor.
The network $F_\theta(z_t,t,c)$ directly predicts the resulting
path-consistent velocity and is trained with
\begin{equation}
\mathcal L_{\mathrm{FM}}^{\mathcal A}
=
\mathbb E_{(x_0,c),\epsilon,t}
\left[
\left\|
F_\theta(z_t,t,c)-v_t^\star
\right\|_2^2
\right].
\label{eq:scale-aware-fm-loss}
\end{equation}

\paragraph{Relation to standard flow matching.}
When $\mathcal A_t\equiv\mathcal I$, the observation is fixed at
$y_t=x_0$, giving $\dot y_t=0$.
Equations~\eqref{eq:scale-aware-path} and
\eqref{eq:scale-aware-velocity} then reduce to
\begin{equation}
z_t=t x_0+(1-t)\epsilon,
\qquad
v_t^\star=x_0-\epsilon,
\end{equation}
which recovers the standard flow-matching path and velocity target.
Thus, standard flow matching is recovered as the identity-operator
special case of our framework.

\subsection{Gaussian--Lanczos Operator Construction}
\label{sec:gaussian-lanczos}

\paragraph{Kernel construction.}
Let
$\mathcal R=\{-R,\ldots,R\}$
denote the discrete kernel support.
On this support, we define the Lanczos and Gaussian components as
\begin{equation}
\begin{aligned}
h_f[r]
&=
\mathbf 1_{\{|r|<2f\}}
\operatorname{sinc}\!\left(\frac{r}{f}\right)
\operatorname{sinc}\!\left(\frac{r}{2f}\right),\\
g_\sigma[r]
&=
\exp\!\left(-\frac{r^2}{2\sigma^2}\right),
\qquad r\in\mathcal R,
\end{aligned}
\label{eq:gaussian-lanczos-components}
\end{equation}
where
$\operatorname{sinc}(a)=\sin(\pi a)/(\pi a)$
and $\operatorname{sinc}(0)=1$.
For the limiting case $\sigma=0$, we define
$g_0:=\delta_0$, with $\delta_0$ denoting the discrete unit impulse.
We then form the one-dimensional Gaussian--Lanczos kernel by
\begin{equation}
w_t
=
\mathcal N\!\left(
\mathcal N(h_{f(t)})
\ast_R
\mathcal N(g_{\sigma(t)})
\right),
\label{eq:gaussian-lanczos-kernel}
\end{equation}
where $\ast_R$ denotes convolution restricted to the support
$\mathcal R$ and $\mathcal N(\cdot)$ denotes kernel normalization.
The corresponding separable two-dimensional kernel defines the observation
operator as
\begin{equation}
\mathcal A_t=(w_t\otimes w_t)\ast_2,
\label{eq:gaussian-lanczos-operator}
\end{equation}
where $\otimes$ denotes the outer product and $\ast_2$ denotes
channel-wise two-dimensional convolution.
The resulting observation is therefore
\begin{equation}
y_t=(w_t\otimes w_t)\ast_2 x_0.
\end{equation}

\begin{figure}[t]
    \centering
    \includegraphics[width=\columnwidth]{
        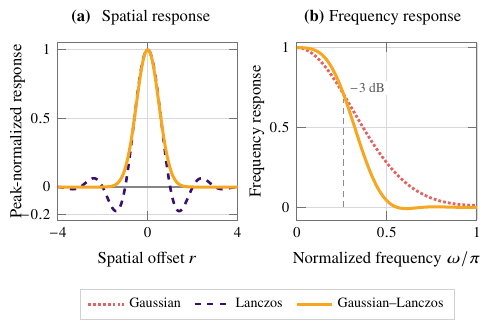
    }
    \caption{\textbf{Spatial and frequency responses of the
    Gaussian--Lanczos kernel.}
    \textit{(a) Spatial response.}
    Gaussian smoothing attenuates the signed sidelobes of the Lanczos
    component, reducing potential ringing.
    \textit{(b) Frequency response.}
    At a matched $-3\,\mathrm{dB}$ cutoff, the Gaussian--Lanczos kernel
    exhibits a narrower transition band than the Gaussian-only kernel,
    yielding sharper scale selectivity and stronger attenuation of
    high-frequency detail.}
    \label{fig:gaussian-lanczos-kernels}
\end{figure}

The Gaussian and Lanczos components provide complementary properties:
Lanczos filtering enables sharper scale selectivity, while Gaussian
smoothing suppresses its ringing artifacts.
As shown in Fig.~\ref{fig:gaussian-lanczos-kernels}(a), Gaussian
smoothing attenuates the oscillatory sidelobes of the Lanczos kernel in
the spatial domain.
In the frequency domain, Fig.~\ref{fig:gaussian-lanczos-kernels}(b)
shows that the combined kernel has a narrower transition band than
Gaussian smoothing alone at the same $-3\,\mathrm{dB}$ cutoff, enabling
sharper frequency selectivity.

\paragraph{Time-dependent scale schedule.}
We couple the kernel scales to flow time through
\begin{equation}
f(t)
=
1+(f_0-1)(1-t),
\qquad
\sigma(t)
=
\sigma_0(1-t),
\label{eq:gaussian-lanczos-schedule}
\end{equation}
where $f_0>1$ and $\sigma_0>0$.
As $t$ increases, both $f(t)$ and $\sigma(t)$ decrease toward their
identity-filtering limits.
The resulting broader kernels at early times suppress fine-scale
variations, while the progressively narrower kernels recover edges and
textures as the path approaches the clean image.

At $t=1$, the Gaussian component is defined by its discrete limit
$g_0=\delta_0$.
Moreover, $f(1)=1$, for which the sampled Lanczos kernel satisfies
$h_1=\delta_0$ on the integer grid.
Hence, the endpoint kernel is obtained by continuous extension:
\begin{equation}
w_1
:=
\lim_{t\to1^-}w_t
=
\delta_0,
\qquad
y_1
=
\mathcal A_1(x_0)
=
x_0.
\end{equation}
Thus, the observation trajectory continuously transitions from a coarse
low-pass observation to exact clean image.

\paragraph{Path derivative.}
For $t\in(0,1)$, the time derivative of the one-dimensional kernel
follows from the chain rule:
\begin{equation}
\dot w_t
=
\frac{\partial w_t}{\partial f}\dot f(t)
+
\frac{\partial w_t}{\partial \sigma}\dot\sigma(t).
\label{eq:kernel-time-derivative}
\end{equation}
Differentiating the corresponding separable two-dimensional kernel
then gives
\begin{equation}
\dot y_t
=
\left(
\dot w_t\otimes w_t
+
w_t\otimes\dot w_t
\right)\ast_2x_0.
\label{eq:operator-time-derivative}
\end{equation}
Because the Lanczos kernel is piecewise smooth due to its
finite-support indicator, the derivative with respect to $t$ exists
almost everywhere, which is sufficient for the flow-matching objective.
The resulting $\dot y_t$ induces the additional $t\dot y_t$ term in the
path-consistent velocity target.

\subsection{Coarse-to-Fine Gaussian--Lanczos Decoder}
\label{sec:backbone}

Pixel-space generation requires both global context modeling and
spatially precise refinement.
Following the two-stage Patch--Pixel design of PixelDiT, our backbone
comprises a Patch-DiT encoder and a GL-CoDA decoder.
The Patch-DiT encoder captures global semantics at the patch level,
while the GL-CoDA decoder progressively refines the representation at
the pixel level.
Each GL-CoDA block contains an attention branch and a parallel
\emph{GL structure branch}.
The former models long-range dependencies over a linearly compressed
set of pixel tokens, while the latter injects Gaussian--Lanczos
responses that progress from coarse to fine across decoder depth.
This design extends the same scale-aware principle from the temporal
evolution of supervision to the spatial refinement of network features.

Concretely, given the normalized pixel features $h_l$ at the $l$-th
GL-CoDA block, we first rearrange the pixel tokens into a full
2D feature map to restore their spatial adjacency.
We then apply $2\times$ average pooling to suppress local noise and
reduce computational cost, obtaining a spatially compressed feature
map $F_l$.
The structural operators reuse the Gaussian--Lanczos construction
defined in Sec.~\ref{sec:gaussian-lanczos}.
We assign four scales $\{n_1,\dots,n_4\}$ to successive decoder blocks
in coarse-to-fine order, with the finest-scale operator
$\mathcal A_{n_4}=\mathcal I$.
The structural response at the $l$-th block is
\begin{equation}
E_l =
\begin{cases}
\mathcal A_{n_1}(F_l), & l=1,\\[2pt]
\mathcal A_{n_l}(F_l)-\mathcal A_{n_{l-1}}(F_l), & l=2,3,4.
\end{cases}
\label{eq:depth-aligned-structure}
\end{equation}

The first block captures the coarse-scale structure, while subsequent
blocks extract progressively finer structural increments by taking
differences between adjacent operator responses.
This adjacent-difference construction prevents redundant re-injection
of coarse structures across decoder depths, allowing each block to
provide complementary information at its designated scale.
Each response is then refined by a lightweight adapter $\mathcal P_l$
and injected into the pixel representation produced by the attention
branch:
\begin{equation}
\widetilde h_l = h_l^{\mathrm{attn}} + \mathcal P_l(E_l),
\label{eq:structure-injection}
\end{equation}
where $\mathcal P_l$ upsamples the response by $2\times$, reversing the
earlier pooling operation, and rearranges it into the pixel-token space.
The resulting residual injection preserves the full-resolution main
stream while progressively enriching its local structure from coarse
to fine.

\paragraph{Training Objective.}
We train the model with the proposed path-consistent flow-matching
objective together with REPA representation alignment
\begin{equation}
\mathcal{L}
=\mathcal L_{\mathrm{FM}}^{\mathcal A}
+\mathcal{L}_{\mathrm{REPA}}.
\label{eq:training-objective}
\end{equation}
In practice, we use direct $v$ prediction and evaluate $\mathcal L_{\mathrm{FM}}^{\mathcal A}$ in $v$ space due to its good optimization stability and generation performance.

\begin{table*}[t]
    \footnotesize
    \centering
    \setlength{\tabcolsep}{4.5pt}
    \renewcommand{\arraystretch}{1.04}
    \resizebox{0.9\linewidth}{!}{%
    \begin{tabular}{@{}clccccccc@{}}
    \toprule
    \textbf{Space} & \textbf{Method} & \textbf{Epochs} & \textbf{\#params}
    & \textbf{gFID}$\downarrow$ & \textbf{sFID}$\downarrow$
    & \textbf{IS}$\uparrow$ & \textbf{Precision}$\uparrow$
    & \textbf{Recall}$\uparrow$ \\
    \midrule
    \multirow{4}{*}{Latent}
    & REPA~\citep{yu2025repa} & 800 & 675M & 1.42 & 4.70 & 305.7 & 0.80 & 0.65 \\
    & SVG-XL~\citep{shi2025svg} & 1400 & 675M & 1.92 & -- & 264.9 & -- & -- \\
    & DDT-XL~\citep{wang2025ddt} & 400 & 675M & 1.26 & -- & 310.6 & 0.79 & 0.65 \\
    & RAE-XL~\citep{zheng2025rae} & 800 & 839M & \textbf{1.13} & -- & 262.6 & 0.78 & 0.67 \\
    \midrule
    \multirow{14}{*}{Pixel}
    & FractalMAR-H~\citep{li2025fractal} & 600 & 844M & 6.15 & -- & 348.9 & 0.81 & 0.46 \\
    & FARMER~\citep{zheng2026farmer} & 320 & 1.9B & 3.60 & -- & 269.2 & 0.81 & 0.51 \\
    & EPG-XXL/16~\citep{lei2026epg} & 600 & 789M & 1.81 & -- & 294.6 & 0.80 & 0.61 \\
    & DeCo-XL/16~\citep{ma2026deco} & 800 & 682M & 1.62 & 4.41 & 301.0 & 0.80 & 0.62 \\
    & PixelFlow-XL~\citep{chen2025pixelflow} & 320 & 677M & 1.98 & 5.83 & 282.1 & 0.81 & 0.60 \\
    & PixNerd-XL~\citep{wang2026pixnerd} & 320 & 700M & 1.93 & -- & 298.0 & 0.80 & 0.60 \\
    & JiT-G~\citep{li2025back} & 600 & 2B & 1.82 & -- & 292.6 & 0.79 & 0.62 \\
    & DiP-XL/16~\citep{chen2026dip} & 600 & 631M & 1.79 & 4.59 & 281.9 & 0.80 & 0.63 \\
    & PixelGen-XL/16~\citep{ma2026pixelgen} & 160 & 676M & 1.83 & -- & 293.6 & 0.79 & 0.63 \\
    & PixelREPA-H/16~\citep{shin2026pixelrepa} & 600 & 953M & 1.81 & -- & 317.2 & -- & -- \\
    & FREPix-XL~\citep{frepix} & 320 & 674M & 1.91 & -- & 295.6 & 0.79 & 0.62 \\
    & PixelDiT-XL~\citep{yu2026pixeldit} & 320 & 797M & \underline{1.61} & 4.68 & 292.7 & 0.78 & 0.64 \\
    \cmidrule(l){2-9}
    \rowcolor{tableOurs}
    & \textbf{Ours} & 80 & 798M & 1.95 & 4.82 & 283.2 & 0.78 & 0.62 \\
    \rowcolor{tableOurs}
    & \textbf{Ours} & 260 & 798M & \textbf{1.52} & 4.61 & 300.0 & 0.78 & 0.64 \\
    \bottomrule
    \end{tabular}%
    }
    \caption{\textbf{Quantitative results} on ImageNet
    $256\times256$ for class-conditioned generation.}
    \vspace{-3mm}
    \label{tab:imagenet256-results}
\end{table*}

\section{Experiments}
\label{sec:experiments}

\begin{figure}[ht]
    \centering
    \includegraphics[width=\columnwidth]
    {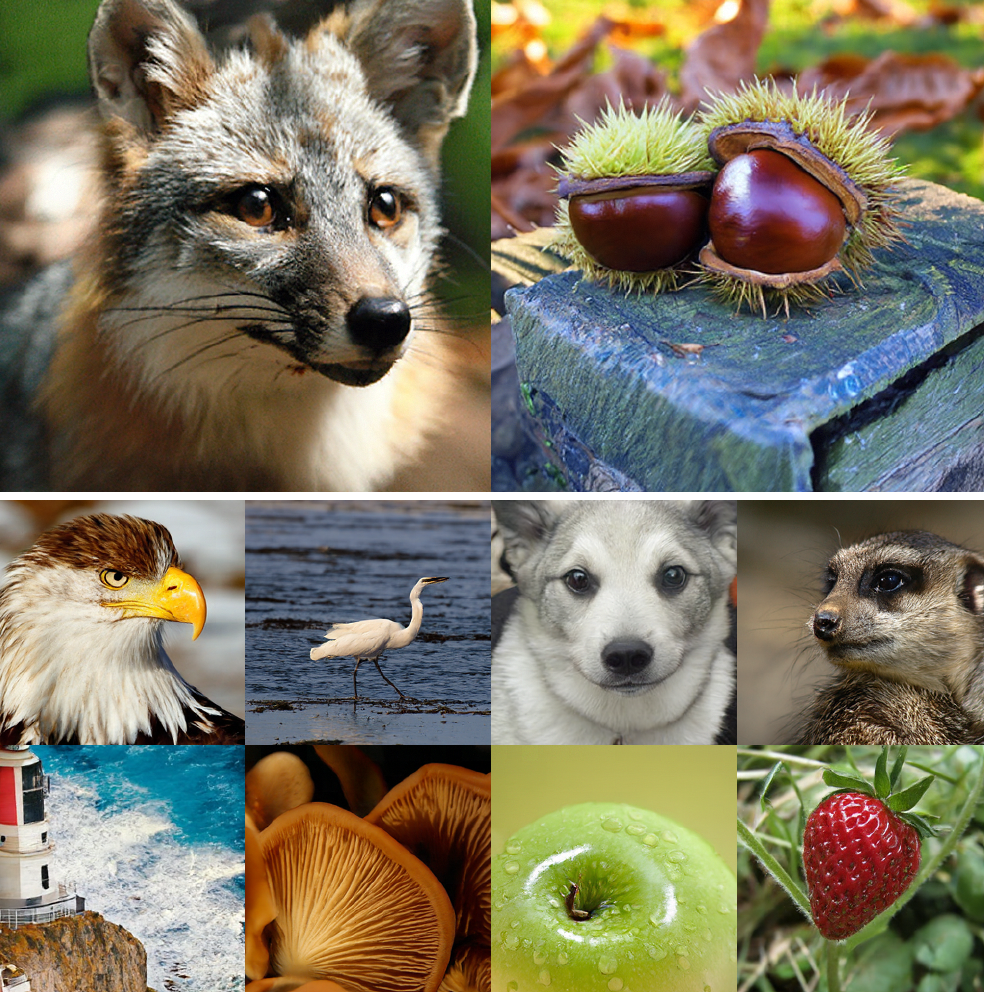}
        \vspace{-5mm}
    \caption{\textbf{Class-conditional samples on ImageNet.}
    Top: samples at $512\times512$. Bottom: samples at $256\times256$.}
    \vspace{-5mm}
    \label{fig:qualitative-results}
\end{figure}

\begin{table}[t]
    \footnotesize
    \centering
    \setlength{\tabcolsep}{3pt}
    \renewcommand{\arraystretch}{1.04}
    \resizebox{\linewidth}{!}{%
    \begin{tabular}{@{}clccccc@{}}
    \toprule
    \textbf{Space} & \textbf{Method}
    & \textbf{gFID}$\downarrow$ & \textbf{sFID}$\downarrow$
    & \textbf{IS}$\uparrow$ & \textbf{Precision}$\uparrow$
    & \textbf{Recall}$\uparrow$ \\
    \midrule
    \multirow{6}{*}{Latent}
    & DiT-XL~\citep{peebles2023dit} & 3.04 & 5.02 & 240.8 & 0.84 & 0.54 \\
    & SiT-XL~\citep{ma2024sit} & 2.62 & 4.18 & 252.2 & 0.84 & 0.57 \\
    & MaskDiT~\citep{zheng2023maskdit} & 2.50 & 5.10 & 256.2 & 0.83 & 0.56 \\
    & U-ViT-H~\citep{bao2023uvit} & 4.05 & -- & 263.8 & 0.84 & 0.48 \\
    & REPA~\citep{yu2025repa} & 2.08 & 4.19 & 274.6 & 0.83 & 0.58 \\
    & RAE-XL~\citep{zheng2025rae} & 1.13 & -- & 259.6 & 0.80 & 0.63 \\
    \midrule
    \multirow{8}{*}{Pixel}
    & PixNerd-XL~\citep{wang2026pixnerd} & 2.84 & 5.95 & 245.6 & 0.80 & 0.59 \\
    & EPG-L/32~\citep{lei2026epg} & 2.35 & -- & 295.4 & 0.82 & 0.57 \\
    & JiT-H~\citep{li2025back} & 1.94 & -- & 309.1 & -- & -- \\
    & PixelDiT-XL~\citep{yu2026pixeldit} & 1.81 & 5.61 & 278.6 & 0.78 & 0.67 \\
    & DiP-XL/32~\citep{chen2026dip} & 2.31 & 4.48 & 291.7 & 0.84 & 0.58 \\
    & DeCo-XL/16~\citep{ma2026deco} & 2.22 & 4.67 & 290.0 & 0.80 & 0.60 \\
    & PixelU-H/32~\citep{guo2026pixelu} & 1.92 & 5.98 & 322.1 & 0.80 & 0.58 \\
    \cmidrule(l){2-7}
    \rowcolor{tableOurs}
    & \textbf{Ours} & \textbf{1.58} & 4.86 & 299.8 & 0.79 & 0.69 \\
    \bottomrule
    \end{tabular}%
    }
    \vspace{-1mm}
    \caption{\textbf{Quantitative comparison} on ImageNet
    $512\times512$.}
    \vspace{-5mm}
    \label{tab:imagenet512-results}
\end{table}

\paragraph{Evaluation setup.}
We conduct class-conditional generation experiments on ImageNet-1K
\citep{deng2009imagenet} at $256^2$ and $512^2$ resolutions.
Unless otherwise specified, all models and ablations are trained for 80 epochs.
We evaluate generation quality using 50K generated samples and report
Fr\'echet Inception Distance (FID)~\citep{heusel2017gans}, spatial FID
(sFID), Inception Score (IS), precision, and recall.

\paragraph{Architecture and Implementation Details.}
Following the two-stage Patch--Pixel design and XL/16 configuration of
PixelDiT~\citep{yu2026pixeldit}, our backbone comprises a 26-block
Patch-DiT encoder with hidden dimension $1152$ and a four-block GL-CoDA
decoder.
For the Gaussian--Lanczos (G--L) observation path, we set $f_0=2$ and
$\sigma_0=1$ with a fixed kernel support of 15 taps.
The kernel scales are linearly annealed over flow time, and the
corresponding analytic derivative is used to construct the velocity target.

\paragraph{Quantitative Results.}
Tables~\ref{tab:imagenet256-results} and~\ref{tab:imagenet512-results} compare our method with existing class-conditional image generators.
On ImageNet $256\times256$, our model reaches a gFID of $1.95$ after only 80 training epochs. 
With extended training to 260 epochs, the gFID further improves to
$1.52$, establishing a new state of the art among pixel-space models
while using 60 fewer training epochs than the previous best.
Metrics such as sFID and IS further validate this improvement, indicating that the gain in fidelity does not come at the expense of sample diversity.

At $512\times512$, our model achieves a gFID of $1.58$, improving upon the previous best pixel-space result of $1.81$ by $12.7\%$.
Recall also increases from $0.67$ to $0.69$, suggesting improved
distributional coverage alongside the lower gFID.
This performance at four times the pixel count demonstrates the scalability of the proposed observation path.

\begin{table}[t]
\centering
\footnotesize
\setlength{\tabcolsep}{3.5pt}
\renewcommand{\arraystretch}{1.08}
\begin{tabular}{@{}
    >{\raggedright\arraybackslash}p{0.22\columnwidth}
    >{\centering\arraybackslash}p{0.15\columnwidth}
    >{\centering\arraybackslash}p{0.27\columnwidth}
    >{\centering\arraybackslash}p{0.27\columnwidth}@{}}
\toprule
Architecture & \#params & Full-image path & G--L path \\
\midrule
JiT
& 2B
& 4.11
& \textbf{3.79} \\
PixelDiT
& 797M
& 2.36
& \textbf{2.14} \\
\rowcolor{tableOurs}
\textbf{Ours}
& 798M
& 2.22
& \textbf{1.95} \\
\bottomrule
\end{tabular}
\caption{\textbf{Observation-Path Ablation across Architectures, measured by FID.}}
\label{tab:path-architecture-ablation}
\end{table}

\paragraph{Qualitative Results.}
Figure~\ref{fig:qualitative-results} shows representative
class-conditional samples at both resolutions. 
At $256\times256$, our method yields category-faithful shapes and coherent layouts while
preserving fine structures such as mushroom gills and water droplets.
At $512\times512$, it further resolves high-frequency detail such as
animal fur and weathered wood while maintaining global consistency.
Together, these results show that our coarse-to-fine
design reconciles semantic coherence with fine-detail fidelity across
resolutions.

\subsection{Ablation Study and Analysis} 
\paragraph{Depth-wise Structure Injection.}
Figure~\ref{fig:depth-feature-visualization} visualizes the input features of the GL-CoDA blocks and the corresponding structural responses produced by the GL structure branch.
At \(t=0.25\), the path state is heavily corrupted by noise, 
leaving object structures largely obscured in the block inputs.
In contrast, the extracted responses exhibit a clear coarse-to-fine
progression across depth: shallow blocks capture smooth global
contours, while deeper blocks progressively highlight object
boundaries and fine details. 
This behavior demonstrates that the GL structure branch extracts
scale-specific structural cues from noisy intermediate
representations and injects them progressively into the main feature
stream.

Table~\ref{tab:path-architecture-ablation} compares the standard
full-image flow-matching path with the proposed Gaussian--Lanczos
observation path across three architectures.
The proposed path consistently reduces FID across all architectures,
demonstrating that the benefit of scale-adaptive supervision is not
specific to a particular backbone.
Our Patch--Pixel backbone with GL-CoDA achieves the best performance,
suggesting that time-dependent target refinement and depth-wise
structural refinement provide complementary benefits.

\begin{table}[t]
\centering
\footnotesize
\setlength{\tabcolsep}{3.5pt}
\renewcommand{\arraystretch}{1.08}
\begin{tabular}{@{}
    >{\raggedright\arraybackslash}p{0.22\columnwidth}
    >{\centering\arraybackslash}p{0.13\columnwidth}
    >{\raggedright\arraybackslash}p{0.40\columnwidth}
    >{\centering\arraybackslash}p{0.16\columnwidth}@{}}
\toprule
\multicolumn{2}{c}{(a) Kernel} &
\multicolumn{2}{c}{(b) Depth assignment} \\
\cmidrule(r){1-2}\cmidrule(l){3-4}
Kernel & FID$\downarrow$ & Assignment & FID$\downarrow$ \\
\midrule
Identity & 2.22 & Coarse only & 2.28 \\
Gaussian & 2.10 & Fine only & 2.12 \\
Lanczos & 2.08 & Fine $\rightarrow$ Coarse & 2.30 \\
\rowcolor{tableOurs}
\textbf{G--L} & \textbf{1.95} &
\textbf{Coarse $\rightarrow$ Fine} & \textbf{1.95} \\
\bottomrule
\end{tabular}
\caption{\textbf{Kernel-Component and Depth-Assignment Ablations.}}
\vspace{-5mm}
\label{tab:kernel-ablation}
\label{tab:depth-operator-ablation}
\end{table}

\paragraph{Kernel Design and Depth Assignment.}
Table~\ref{tab:kernel-ablation}(a) ablates the components of the
observation kernel.
Both Gaussian and Lanczos filtering improve FID over the baseline,
while their combination performs best, reducing FID to 1.95.
This improvement reflects the complementary properties of the two
operators: Lanczos provides sharper scale selectivity, while Gaussian
smoothing attenuates its oscillatory sidelobes and reduces ringing
artifacts, yielding a sharper yet cleaner coarse-scale observation.

Table~\ref{tab:depth-operator-ablation}(b) further shows that the
injection order is critical.
Using only coarse or only fine responses is inferior to the joint
configuration, while reversing their order leads to a further
performance drop.
The best FID is obtained when coarse responses are assigned to shallow
blocks and progressively finer responses to deeper blocks.
These results show that the benefit comes not merely from introducing
the structural operator, but from matching its coarse-to-fine ordering
with the progression of decoder depth.

\begin{table}[t]
\centering
\footnotesize
\setlength{\tabcolsep}{3.5pt}
\renewcommand{\arraystretch}{1.08}
\begin{tabular}{@{}
    >{\centering\arraybackslash}p{0.21\columnwidth}
    >{\centering\arraybackslash}p{0.21\columnwidth}
    >{\centering\arraybackslash}p{0.29\columnwidth}
    >{\centering\arraybackslash}p{0.17\columnwidth}@{}}
\toprule
\multicolumn{2}{c}{Initial kernel scales}
& \multirow{2}{*}{Schedule}
& \multirow{2}{*}{FID$\downarrow$} \\
\cmidrule(lr){1-2}
Lanczos $f_0$ & Gaussian $\sigma_0$ & & \\
\midrule
$4$ & $1$ & Linear & 2.00 \\
$2$ & $2$ & Linear & 2.06 \\
\rowcolor{tableOurs}
$2$ & $1$ & Linear & \textbf{1.95} \\
$2$ & $1$ & Cosine & 2.08 \\
\bottomrule
\end{tabular}
\caption{\textbf{Kernel-Scale and Temporal-Schedule Sensitivity.}}
\label{tab:schedule-ablation}
\end{table}

\begin{figure}[t]
    \centering
    \setlength{\tabcolsep}{0pt}
    \renewcommand{\arraystretch}{1.0}
    \begin{tabular}{@{}c@{\hspace{2.5pt}}cccc@{}}
        \includegraphics[width=0.198\columnwidth]{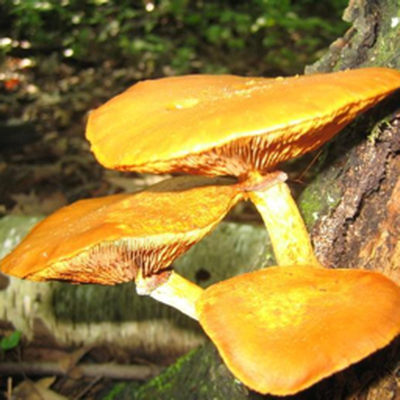} &
        \includegraphics[width=0.198\columnwidth]{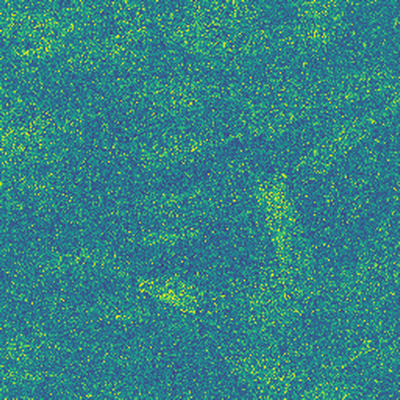} &
        \includegraphics[width=0.198\columnwidth]{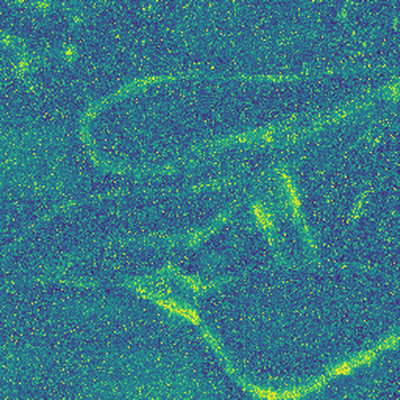} &
        \includegraphics[width=0.198\columnwidth]{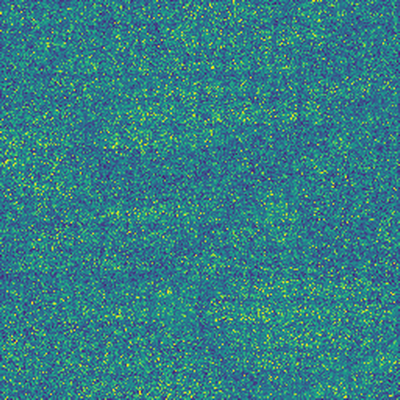} &
        \includegraphics[width=0.198\columnwidth]{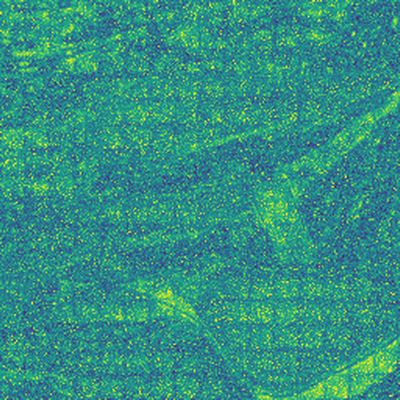} \\
        \includegraphics[width=0.198\columnwidth]{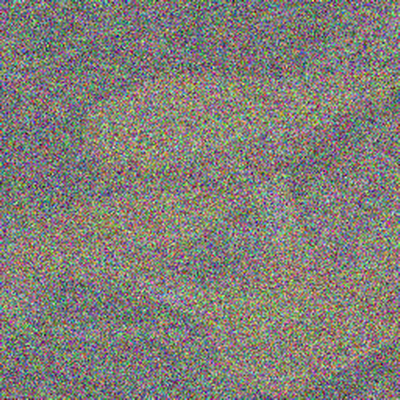} &
        \includegraphics[width=0.198\columnwidth]{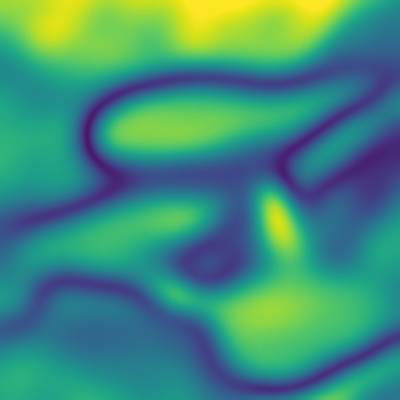} &
        \includegraphics[width=0.198\columnwidth]{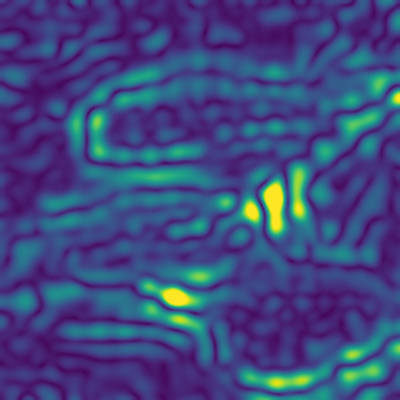} &
        \includegraphics[width=0.198\columnwidth]{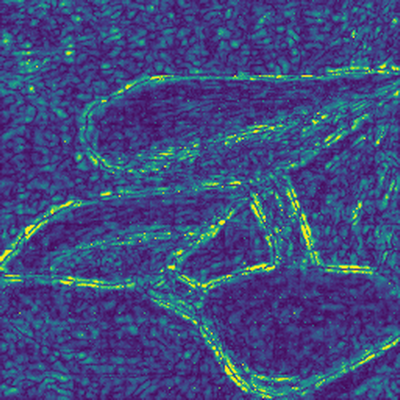} &
        \includegraphics[width=0.198\columnwidth]{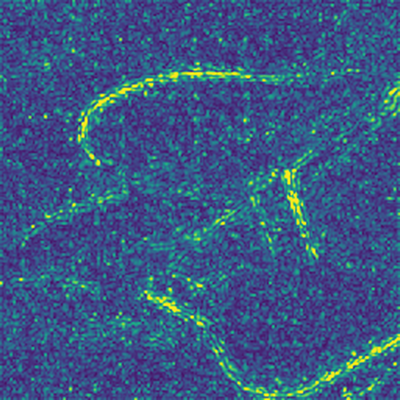} \\
        \footnotesize $x_0/z_{0.25}$ &
        \footnotesize B1 & \footnotesize B2 & \footnotesize B3 & \footnotesize B4
    \end{tabular}
    \caption{\textbf{Depth-wise feature visualization at $t=0.25$.}
    The first column shows $x_0$ (top) and $z_{0.25}$ (bottom);
    B1--B4 show block input features (top) and our injected
    Gaussian--Lanczos structural responses (bottom).}
    \vspace{-3mm}
    \label{fig:depth-feature-visualization}
\end{figure}

\paragraph{Sensitivity to Scale and Schedule.}
Table~\ref{tab:schedule-ablation} shows that, within the tested range,
the model is robust to $f_0$ and the schedule, while being more
sensitive to $\sigma_0$.
Our final configuration of $f_0=2$, $\sigma_0=1$, and a linear schedule
achieves the best FID of $1.95$.

\section{Conclusion}
In this paper, we identify a scale--time mismatch as an overlooked
source of optimization difficulty in pixel-space diffusion:
image structures follow a coarse-to-fine predictability order as
denoising proceeds, whereas existing models supervise all spatial
scales at every step, producing low-SNR gradients that slow convergence.
To address this, we propose an observation-operator diffusion
framework that replaces fixed full-image supervision with a
time-indexed observation trajectory, together with a path-consistent
training objective.
A family of Gaussian--Lanczos operators provides scale-specific
observations that progressively transition from coarse structures to
fine details.
We further deploy these operators across decoder depth through the
GL-CoDA decoder, extending the same coarse-to-fine principle from
diffusion time to network depth.
On ImageNet-256, our method converges faster and achieves an FID of
$1.52$, showing that aligning the supervised scale with the
structure reliably recoverable at each denoising step improves
pixel-space diffusion.
Our experiments focus on natural images; extending the proposed
scale-aware framework to specialized domains such as medical imaging
is a promising direction for future work.

\bibliography{aaai2027}

@inproceedings{rombach2022high,
  author    = {Rombach, Robin and Blattmann, Andreas and Lorenz, Dominik and Esser, Patrick and Ommer, Bjorn},
  title     = {High-Resolution Image Synthesis with Latent Diffusion Models},
  booktitle = {Proceedings of the IEEE/CVF Conference on Computer Vision and Pattern Recognition},
  pages     = {10684--10695},
  year      = {2022}
}

@misc{li2025back,
  author        = {Li, Tianhong and He, Kaiming},
  title         = {Back to Basics: Let Denoising Generative Models Denoise},
  year          = {2025},
  eprint        = {2511.13720},
  archivePrefix = {arXiv},
  primaryClass  = {cs.CV},
  url           = {https://arxiv.org/abs/2511.13720}
}

@misc{chen2025pixelflow,
  author        = {Chen, Shoufa and Ge, Chongjian and Zhang, Shilong and Sun, Peize and Luo, Ping},
  title         = {{PixelFlow}: Pixel-Space Generative Models with Flow},
  year          = {2025},
  eprint        = {2504.07963},
  archivePrefix = {arXiv},
  primaryClass  = {cs.CV},
  url           = {https://arxiv.org/abs/2504.07963}
}

@inproceedings{yu2026pixeldit,
  author    = {Yu, Yongsheng and Xiong, Wei and Nie, Weili and Sheng, Yichen and Liu, Shiqiu and Luo, Jiebo},
  title     = {{PixelDiT}: Pixel Diffusion Transformers for Image Generation},
  booktitle = {Proceedings of the IEEE/CVF Conference on Computer Vision and Pattern Recognition},
  pages     = {14273--14282},
  year      = {2026}
}

@misc{chen2026asymflow,
  author        = {Chen, Hansheng and Ackermann, Jan and Kim, Minseo and Wetzstein, Gordon and Guibas, Leonidas},
  title         = {Asymmetric Flow Models},
  year          = {2026},
  eprint        = {2605.12964},
  archivePrefix = {arXiv},
  primaryClass  = {cs.CV},
  url           = {https://arxiv.org/abs/2605.12964}
}

@misc{jin2026revisiting,
  author        = {Jin, Qing and Wang, Chaoyang},
  title         = {Revisiting Diffusion Model Predictions Through Dimensionality},
  year          = {2026},
  eprint        = {2601.21419},
  archivePrefix = {arXiv},
  primaryClass  = {cs.CV},
  url           = {https://arxiv.org/abs/2601.21419}
}

@inproceedings{lipman2023flow,
  author    = {Lipman, Yaron and Chen, Ricky T. Q. and Ben-Hamu, Heli and Nickel, Maximilian and Le, Matt},
  title     = {Flow Matching for Generative Modeling},
  booktitle = {International Conference on Learning Representations},
  year      = {2023},
  url       = {https://openreview.net/forum?id=PqvMRDCJT9t}
}

@article{hoogeboom2022blurring,
  title={Blurring diffusion models},
  author={Hoogeboom, Emiel and Salimans, Tim},
  journal={arXiv preprint arXiv:2209.05557},
  year={2022}
}

@inproceedings{peebles2023dit,
  author    = {Peebles, William and Xie, Saining},
  title     = {Scalable Diffusion Models with Transformers},
  booktitle = {Proceedings of the IEEE/CVF International Conference on Computer Vision},
  pages     = {4195--4205},
  year      = {2023}
}

@inproceedings{ma2024sit,
  author    = {Ma, Nanye and Goldstein, Mark and Albergo, Michael S. and Boffi, Nicholas M. and Vanden-Eijnden, Eric and Xie, Saining},
  title     = {{SiT}: Exploring Flow and Diffusion-Based Generative Models with Scalable Interpolant Transformers},
  booktitle = {European Conference on Computer Vision},
  year      = {2024}
}

@article{zheng2023maskdit,
  author  = {Zheng, Hongkai and Nie, Weili and Vahdat, Arash and Anandkumar, Anima},
  title   = {Fast Training of Diffusion Models with Masked Transformers},
  journal = {Transactions on Machine Learning Research},
  year    = {2023}
}

@article{shi2025svg,
  author  = {Shi, Minglei and Wang, Haolin and Zheng, Wenzhao and Yuan, Ziyang and Wu, Xiaoshi and Wang, Xintao and Wan, Pengfei and Zhou, Jie and Lu, Jiwen},
  title   = {Latent Diffusion Model without Variational Autoencoder},
  journal = {arXiv preprint arXiv:2510.15301},
  year    = {2025}
}

@misc{wang2025ddt,
  author        = {Wang, Shuai and Tian, Zhi and Huang, Weilin and Wang, Limin},
  title         = {{DDT}: Decoupled Diffusion Transformer},
  year          = {2025},
  eprint        = {2504.05741},
  archivePrefix = {arXiv},
  primaryClass  = {cs.CV}
}

@article{zheng2025rae,
  author  = {Zheng, Boyang and Ma, Nanye and Tong, Shengbang and Xie, Saining},
  title   = {Diffusion Transformers with Representation Autoencoders},
  journal = {arXiv preprint arXiv:2510.11690},
  year    = {2025}
}

@inproceedings{bao2023uvit,
  author    = {Bao, Fan and Nie, Shen and Xue, Kaiwen and Cao, Yue and Li, Chongxuan and Su, Hang and Zhu, Jun},
  title     = {All are Worth Words: A {ViT} Backbone for Diffusion Models},
  booktitle = {Proceedings of the IEEE/CVF Conference on Computer Vision and Pattern Recognition},
  pages     = {22669--22679},
  year      = {2023}
}

@misc{yu2025repa,
  author        = {Yu, Sihyun and Kwak, Sangkyung and Jang, Huiwon and Jeong, Jongheon and Huang, Jonathan and Shin, Jinwoo and Xie, Saining},
  title         = {Representation Alignment for Generation: Training Diffusion Transformers Is Easier Than You Think},
  year          = {2024},
  eprint        = {2410.06940},
  archivePrefix = {arXiv},
  primaryClass  = {cs.CV},
  url           = {https://arxiv.org/abs/2410.06940}
}

@inproceedings{deng2009imagenet,
  author    = {Deng, Jia and Dong, Wei and Socher, Richard and Li, Li-Jia and Li, Kai and Fei-Fei, Li},
  title     = {{ImageNet}: A Large-Scale Hierarchical Image Database},
  booktitle = {Proceedings of the IEEE Conference on Computer Vision and Pattern Recognition},
  pages     = {248--255},
  year      = {2009}
}

@inproceedings{heusel2017gans,
  author    = {Heusel, Martin and Ramsauer, Hubert and Unterthiner, Thomas and Nessler, Bernhard and Hochreiter, Sepp},
  title     = {{GANs} Trained by a Two Time-Scale Update Rule Converge to a Local {Nash} Equilibrium},
  booktitle = {Advances in Neural Information Processing Systems},
  volume    = {30},
  year      = {2017}
}

@inproceedings{dhariwal2021diffusion,
  author    = {Dhariwal, Prafulla and Nichol, Alexander},
  title     = {Diffusion Models Beat {GANs} on Image Synthesis},
  booktitle = {Advances in Neural Information Processing Systems},
  volume    = {34},
  pages     = {8780--8794},
  year      = {2021}
}

@misc{li2025fractal,
  author        = {Li, Tianhong and Sun, Qinyi and Fan, Lijie and He, Kaiming},
  title         = {Fractal Generative Models},
  year          = {2025},
  eprint        = {2502.17437},
  archivePrefix = {arXiv},
  primaryClass  = {cs.LG},
  url           = {https://arxiv.org/abs/2502.17437}
}

@inproceedings{zheng2026farmer,
  author    = {Zheng, Guangting and Zhao, Qinyu and Yang, Tao and Xiao, Fei and Lin, Zhijie and Wu, Jie and Deng, Jiajun and Zhang, Yanyong and Zhu, Rui},
  title     = {{FARMER}: Flow AutoRegressive Transformer over Pixels},
  booktitle = {Proceedings of the IEEE/CVF Conference on Computer Vision and Pattern Recognition},
  pages     = {25730--25741},
  year      = {2026}
}

@inproceedings{lei2026epg,
  author    = {Lei, Jiachen and Liu, Keli and Berner, Julius and Yu, Haiming and Zheng, Hongkai and Wu, Jiahong and Chu, Xiangxiang},
  title     = {There Is No {VAE}: End-to-End Pixel-Space Generative Modeling via Self-Supervised Pre-Training},
  booktitle = {International Conference on Learning Representations},
  year      = {2026}
}

@book{lindeberg1994scale,
  author    = {Lindeberg, Tony},
  title     = {Scale-Space Theory in Computer Vision},
  series    = {The Springer International Series in Engineering and Computer Science},
  volume    = {256},
  publisher = {Kluwer Academic Publishers},
  address   = {Boston, MA},
  year      = {1994},
  doi       = {10.1007/978-1-4757-6465-9}
}

@inproceedings{wang2026pixnerd,
  author    = {Wang, Shuai and Gao, Ziteng and Zhu, Chenhui and Huang, Weilin and Wang, Limin},
  title     = {{PixNerd}: Pixel Neural Field Diffusion},
  booktitle = {International Conference on Learning Representations},
  year      = {2026}
}

@inproceedings{chen2026dip,
  author    = {Chen, Zhennan and Zhu, Junwei and Chen, Xu and Zhang, Jiangning and Hu, Xiaobin and Zhao, Hanzhen and Wang, Chengjie and Yang, Jian and Tai, Ying},
  title     = {{DiP}: Taming Diffusion Models in Pixel Space},
  booktitle = {Proceedings of the IEEE/CVF Conference on Computer Vision and Pattern Recognition},
  pages     = {36136--36146},
  year      = {2026}
}

@misc{ma2026pixelgen,
  author        = {Ma, Zehong and Xu, Ruihan and Zhang, Shiliang},
  title         = {{PixelGen}: Improving Pixel Diffusion with Perceptual Supervision},
  year          = {2026},
  eprint        = {2602.02493},
  archivePrefix = {arXiv},
  primaryClass  = {cs.CV},
  url           = {https://arxiv.org/abs/2602.02493}
}

@misc{shin2026pixelrepa,
  author        = {Shin, Jaeyo and Kim, Jiwook and Shim, Hyunjung},
  title         = {Representation Alignment for Just Image Transformers Is Not Easier Than You Think},
  year          = {2026},
  eprint        = {2603.14366},
  archivePrefix = {arXiv},
  primaryClass  = {cs.CV},
  url           = {https://arxiv.org/abs/2603.14366}
}

@misc{guo2026pixelu,
  author        = {Guo, Zipeng and Ma, Lichen and He, Yu and Fu, Xiaolong and Fu, Jingling and Huang, Junshi and Li, Yan},
  title         = {{PixelU}: A U-Shaped Transformer for Efficient End-to-End Pixel Diffusion},
  year          = {2026},
  eprint        = {2606.27760},
  archivePrefix = {arXiv},
  primaryClass  = {cs.CV},
  url           = {https://arxiv.org/abs/2606.27760}
}

@inproceedings{ho2020denoising,
  title={Denoising Diffusion Probabilistic Models},
  author={Ho, Jonathan and Jain, Ajay and Abbeel, Pieter},
  booktitle={NeurIPS},
  year={2020}
}

@article{witkin1983scale,
  title={Scale-space filtering},
  author={Witkin, Andrew P.},
  journal={International Joint Conference on Artificial Intelligence},
  year={1983}
}

@inproceedings{peebles2023scalable,
  title={Scalable Diffusion Models with Transformers},
  author={Peebles, William and Xie, Saining},
  booktitle={ICCV},
  year={2023}
}

@article{burt1983laplacian,
  title={The Laplacian Pyramid as a Compact Image Code},
  author={Burt, Peter J. and Adelson, Edward H.},
  journal={IEEE Transactions on Communications},
  volume={31},
  number={4},
  pages={532--540},
  year={1983},
  publisher={IEEE}
}

@incollection{turkowski1990filters,
  author = {Turkowski, Ken},
  title = {Filters for Common Resampling Tasks},
  booktitle = {Graphics Gems},
  editor = {Glassner, Andrew S.},
  publisher = {Academic Press},
  year = {1990},
  pages = {147--165}
}

@misc{he2026hyperdit,
  author        = {He, Yu and Ma, Lichen and Guo, Zipeng and Shan, Xinyuan and Fu, Jingling and Chen, Dong and Huang, Junshi and Li, Yan},
  title         = {{HyperDiT}: Hyper-Connected Transformers for High-Fidelity Pixel-Space Diffusion},
  year          = {2026},
  eprint        = {2605.15741},
  archivePrefix = {arXiv},
  primaryClass  = {cs.CV},
  url           = {https://arxiv.org/abs/2605.15741}
}

@inproceedings{ma2026deco,
  author    = {Ma, Zehong and Wei, Longhui and Wang, Shuai and Zhang, Shiliang and Tian, Qi},
  title     = {{DeCo}: Frequency-Decoupled Pixel Diffusion for End-to-End Image Generation},
  booktitle = {Proceedings of the IEEE/CVF Conference on Computer Vision and Pattern Recognition},
  pages     = {43600--43610},
  year      = {2026}
}

@misc{frepix,
  author        = {Lin, Mingfeng and Chen, Jiakun and Han, Liang and Nie, Liqiang},
  title         = {{FREPix}: Frequency-Heterogeneous Flow Matching for Pixel-Space Image Generation},
  year          = {2026},
  eprint        = {2605.06421},
  archivePrefix = {arXiv},
  primaryClass  = {cs.CV},
  url           = {https://arxiv.org/abs/2605.06421}
}

@misc{spectralforcing,
  author        = {Fan, Weichen and Diao, Haiwen and Wu, Penghao and Liu, Ziwei},
  title         = {Show the Signal, Hide the Noise: Spectral Forcing for Pixel-Space Diffusion},
  year          = {2026},
  eprint        = {2606.15236},
  archivePrefix = {arXiv},
  primaryClass  = {cs.CV},
  url           = {https://arxiv.org/abs/2606.15236}
}

@misc{ma2026frequencybooster,
  author        = {Ma, Lichen and Guo, Zipeng and He, Yu and Fu, Xiaolong and Liu, Luohang and Fu, Jingling and Huang, Junshi and Li, Yan},
  title         = {{FrequencyBooster}: Full-Frequency Modeling for High-Fidelity Pixel Diffusion},
  year          = {2026},
  eprint        = {2605.17759},
  archivePrefix = {arXiv},
  primaryClass  = {cs.CV},
  url           = {https://arxiv.org/abs/2605.17759}
}

@misc{tong2026egfm,
  author        = {Tong, Haoyang and He, Yu and Li, Fang and Ma, Lichen and Fu, Jingling and Chen, Dong and Chen, Zhen and Huang, Junshi and Cao, Jie},
  title         = {Energy-Guided Flow Matching},
  year          = {2026},
  eprint        = {2608.05811},
  archivePrefix = {arXiv},
  primaryClass  = {cs.CV},
  url           = {https://arxiv.org/abs/2608.05811}
}

\end{document}